\documentclass[10pt,twocolumn,letterpaper]{article}

\usepackage{cvpr} 

\usepackage{multirow}
\usepackage{multicol}
\usepackage{pifont}
\usepackage{subcaption}
\newcommand{\cmark}{\ding{51}}
\newcommand{\xmark}{\ding{55}}

\newcommand{\name}{PVT-GDLA\xspace}

\usepackage[accsupp]{axessibility}  

\definecolor{cvprblue}{rgb}{0.21,0.49,0.74}
\usepackage[pagebackref,breaklinks,colorlinks,allcolors=cvprblue]{hyperref}

\def\paperID{} 
\def\confName{CVPR}
\def\confYear{2025}

\title{Gated Differential Linear Attention: A Linear-Time Decoder for High-Fidelity Medical Segmentation}

\author{Hongbo Zheng$^{1}$\thanks{Equal Contribution.} \quad Afshin Bozorgpour \!$^{2}$\footnotemark[1] \quad Dorit Merhof \!$^{2}$\thanks{Corresponding authors.} \quad Minjia Zhang$^{1}$\footnotemark[2] \\
University of Illinois Urbana-Champaign$^{1}$ \quad University of Regensburg$^{2}$ \\
{\tt\small \{hongboz2, minjiaz\}@illinois.edu \quad \{afshin.bozorgpour, dorit.merhof\}@ur.de}
}

\begin{document}
\maketitle
\begin{abstract}
Medical image segmentation requires models that preserve fine anatomical boundaries while remaining practical for clinical deployment. Transformers capture long-range dependencies but incur quadratic attention cost, whereas CNNs are efficient but less effective at global reasoning. Linear attention offers \(\mathcal{O}(N)\) scaling, but often produces diffuse feature aggregation that weakens boundary-sensitive prediction. We introduce a gated differential linear-attention mixer for medical image segmentation. Its global path, Gated Differential Linear Attention (GDLA), performs differential subtraction between two kernelized attention branches over complementary query/key subspaces to suppress redundant responses, and employs a data-dependent gate for token refinement. A parallel local token-mixing branch with depthwise convolution strengthens neighborhood interactions for better refinement, and the two branches are fused while preserving \(\mathcal{O}(N)\) complexity. When instantiated in a pretrained Pyramid Vision Transformer (PVT)-based encoder--decoder model, \name achieves state-of-the-art results on the evaluated 2D medical segmentation benchmarks spanning CT, MRI, ultrasound, and dermoscopy, with a favorable accuracy--efficiency trade-off over closely related baselines. The code is publicly available at \href{https://github.com/xmindflow/gdla}{https://github.com/xmindflow/gdla}.
\end{abstract}    
\section{Introduction}
\label{sec:intro}

Medical image segmentation is fundamental to screening, diagnosis, and intervention planning in radiology, ultrasound, and dermatology. Reliable systems must capture \emph{global context} to reason over distant structures while preserving \emph{fine boundary fidelity} for small or thin anatomies, all under the tight computational budgets common in clinical environments. Balancing these requirements remains challenging: convolutional neural networks (CNNs)~\citep{ronneberger2015u,huang2020unet,lou2021dc} are efficient and effective at modeling locality but struggle with long-range dependencies, whereas Transformer backbones \citep{oktay2018attention,chen2021transunet,cao2021swinunet,dong2021polyp} capture global interactions at the cost of quadratic attention complexity and greater demands for data and computation.

Linear attention reduces the memory and time complexity of attention from $\mathcal{O}(N^{2})$ to $\mathcal{O}(N)$ by replacing softmax with kernel feature maps and exploiting associativity~\citep{pmlr-v119-katharopoulos20a,tsai2019transformer,shen2021efficient,Cai_2023_ICCV,han2024demystify}. However, its nonnegative and approximately linear similarity often leads to over-smoothed feature aggregation, producing diffuse, low-contrast responses that blur boundaries, a phenomenon referred to as \emph{attention dilution}~\citep{qin-etal-2022-devil}. Recent work has shown that \emph{differential attention} can sharpen focus by subtracting complementary attention responses~\citep{ye2025differential}, while \emph{lightweight gating} can improve selectivity and stability through input-adaptive modulation~\citep{qiu2025gated}. These observations suggest a promising direction for improving linear attention in dense prediction, provided that such refinements can be incorporated without sacrificing linear-time complexity.

To improve linear attention for dense medical segmentation while preserving $\mathcal{O}(N)$ complexity, we propose \emph{Gated Differential Linear Attention} (GDLA), a linear-attention module that combines differential subtraction, lightweight gating, and local token mixing. \!GDLA computes two kernelized attention branches on complementary query/key subspaces and combines them through learnable subtraction to improve feature selectivity, while a head-specific gate further modulates the resulting attention output for adaptive refinement. In parallel, a lightweight depthwise-convolutional local mixer reinforces neighborhood interactions important for boundary-sensitive prediction. 
We integrate the GDLA mixer into the decoding stages of an encoder--decoder segmentation model built on a pretrained Pyramid Vision Transformer (PVT) encoder~\citep{wang2021pyramid}. Across CT, MRI, ultrasound, and dermoscopy benchmarks, our model achieves consistent gains with a favorable accuracy--efficiency trade-off relative to closely related baselines.

\begin{figure*}[t]
    \centering
    \begin{subfigure}[t]{0.48\linewidth}
        \centering
        \includegraphics[width=\linewidth]{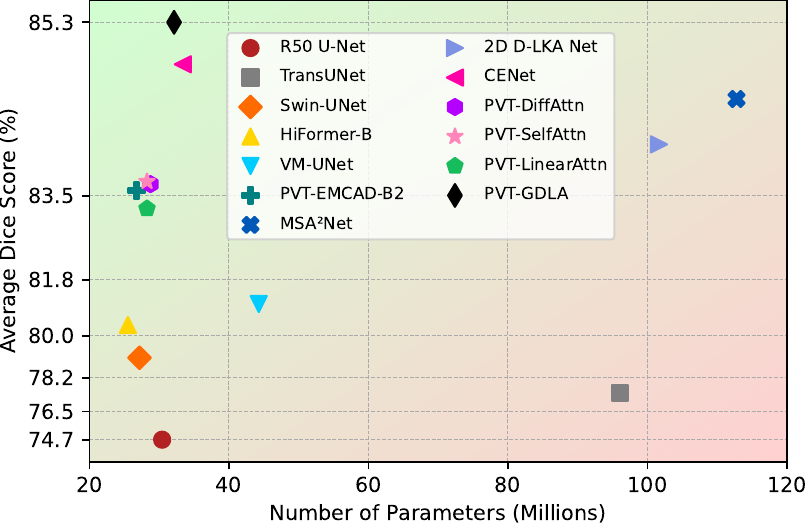}
        \caption{Average Dice score vs. \# of parameters.}
        \label{fig:dice_vs_param}
    \end{subfigure}\hfill
    \begin{subfigure}[t]{0.48\linewidth}
        \centering
        \includegraphics[width=\linewidth]{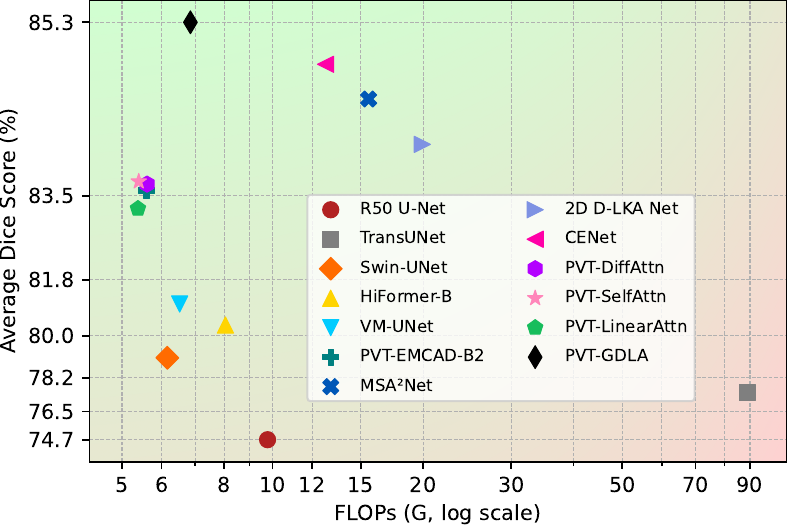}
        \caption{Average Dice score vs. \# of FLOPs.}
        \label{fig:dice_vs_flops}
    \end{subfigure}
    \vspace{-5pt}
    \caption{Comparison of model performance on Synapse dataset and compute--accuracy trade-off. Our approach achieves the highest Dice score with lower \# parameters and lower \# FLOPs.}
    \label{fig:param_flops}
    \vspace{-10pt}
\end{figure*}

Our contributions are as follows:
\ding{182} \textit{GDLA formulation}. We propose Gated Differential Linear Attention, a kernelized linear-attention module that preserves $\mathcal{O}(N)$ complexity while improving selectivity through branch-wise differential subtraction.
\ding{183} \textit{Local refinement design}. We add a lightweight local token-mixing branch and head-wise gating to strengthen neighborhood interactions and boundary recovery with minimal overhead.
\ding{184} \textit{Empirical validation}. Across \textit{Synapse}, \textit{ACDC}, \textit{BUSI}, \textit{HAM10000}, and \textit{PH$^{2}$}, the resulting model consistently improves accuracy over strong linear-attention and transformer baselines with favorable parameter/FLOP trade-offs as shown in Figure~\ref{fig:param_flops}.
\section{Related Work}

\subsection{Medical Segmentation Backbones}

\paragraph{CNN backbones.}
U-shaped CNN architectures remain highly competitive in medical image segmentation because they combine strong locality, translation equivariance, and efficient skip-connected decoding~\citep{ronneberger2015u,lou2021dc,chen2018encoder,huang2020unet}. Many extensions improve multi-scale fusion and boundary recovery through decoder-side context modules, attention mechanisms~\citep{oktay2018attention,wang2022uctransnet}, deformable convolutions~\citep{azad2023selfattention}, or dilated convolutions~\citep{guo2023visual}. Despite their efficiency and strong inductive bias for local structure, purely convolutional models often struggle to capture long-range dependencies that are important for globally coherent predictions in complex anatomies.

\paragraph{Transformer and hybrid backbones.}
Transformer-based encoders and decoders have substantially advanced medical segmentation by improving global context aggregation~\citep{chen2021transunet,cao2021swinunet,dong2021polyp}. However, quadratic self-attention increases memory and computational cost, particularly at high resolution, and often requires larger training sets. Hybrid CNN--Transformer models aim to balance local inductive bias with global reasoning~\citep{chen2021transunet,huang2022missformer}, but this balance remains challenging in dense prediction settings that require both long-range interaction and fine-grained boundary localization.

\subsection{Efficient (Kernelized) Attention}
Quadratic self-attention is effective but computationally expensive for high-resolution dense prediction. Kernelized or linear attention replaces softmax with nonnegative feature maps and exploits associativity to reduce both time and memory complexity from $\mathcal{O}(N^2)$ to $\mathcal{O}(N)$~\citep{pmlr-v119-katharopoulos20a,tsai2019transformer,shen2021efficient,Cai_2023_ICCV,han2024demystify}. This makes linear attention attractive for medical segmentation, where efficiency is important and input resolution can be high. However, the nonnegative structure of kernelized attention can also lead to over-smoothed feature aggregation and diffuse responses, a phenomenon referred to as \emph{attention dilution}~\citep{qin-etal-2022-devil}. Our work is motivated by this trade-off: how to improve the selectivity of linear attention for dense prediction without giving up its linear-time form.

\subsection{Gating Mechanisms in Attention and FFNs}

Gating mechanisms facilitate input-adaptive sparsity and context-aware semantic modulation, enhancing feature representation through nonlinearity. They can selectively amplify relevant features~\citep{fiaz2024guided} while suppressing irrelevant ones~\citep{BozAfs_CENet_MICCAI2025}, improving adaptivity and stability~\citep{li2022moganet}, and mitigating attention sinks with minimal computational overhead~\citep{shazeer2020glu, qiu2025gated}. In practice, post-attention gates and gated feed-forward networks (FFNs) often improve boundary detail, multi-scale representation, and calibration in dense vision tasks, such as medical image segmentation, where highly detailed feature representation is crucial.

\begin{figure*}[th]
    \centering
    \includegraphics[width=1.00\linewidth]{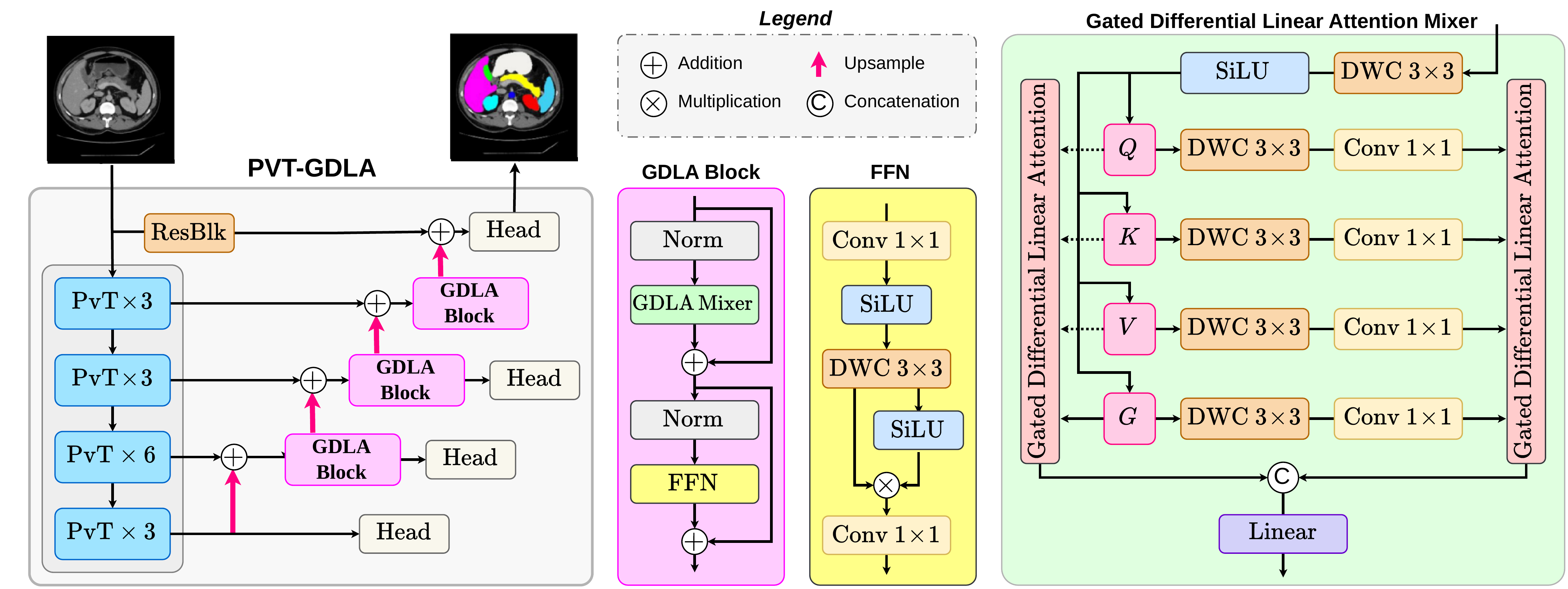}
    \vspace{-12pt}
    \caption{PVT-GDLA overview. A pretrained Pyramid Vision Transformer (PVT) encoder extracts multi-scale features that are progressively decoded by GDLA blocks. Each GDLA block contains a GDLA mixer and a feed-forward network (FFN). Within the mixer, a gated differential linear-attention path captures globally selective context, while a parallel local token-mixing branch reinforces neighborhood interactions. The fused features are projected and upsampled with skip connections to recover spatial resolution.}
    \label{fig:pvtgdla}
    \vspace{-10pt}
\end{figure*}

\subsection{Decoder Design for Medical Segmentation}

Recent decoders emphasize multi-scale fusion and boundary-aware refinement, often combining attention with lightweight convolutions to recover fine detail at modest computational cost~\citep{dong2021polyp,rahman2024emcad,BozAfs_CENet_MICCAI2025}. Specifically, CENet~\citep{BozAfs_CENet_MICCAI2025} augments skip connections with gated context modules and denoising to enhance boundary fidelity across radiology and dermoscopy. EMCAD~\citep{rahman2024emcad} provides an efficient decoding design with multi-scale convolutional decoding, convolutional attention, and a large-kernel grouped attention gate. These results highlight the importance of decoder-side refinement, but such designs remain largely convolution-driven and are less effective at modeling long-range interactions.
\section{Methodology}

We propose the \emph{Gated Differential Linear Attention} (GDLA) mixer, a linear-attention module for dense medical image segmentation. GDLA mixer builds on kernelized linear attention, which offers $\mathcal{O}(N)$ complexity with respect to the number of tokens $N$. We integrate GDLA into the decoder of an encoder--decoder segmentation model with a pretrained Pyramid Vision Transformer (PVT) encoder. The resulting GDLA mixer contains three parts: (i) a \emph{gated differential linear-attention} path for globally selective feature aggregation, (ii) a parallel \emph{local token-mixing} branch for short-range spatial refinement, and (iii) an \emph{output-fusion} step that combines the two. Together, these components preserve the $\mathcal{O}(N)$ complexity of linear attention while improving both global context modeling and local feature refinement. An overview is shown in Figure~\ref{fig:pvtgdla}.

\subsection{Preliminaries}

\subsubsection{Multi-Head Differential Attention} 
\label{subsubse:diff-attn}
Differential attention~\citep{ye2025differential} sharpens focus by subtracting two attention maps computed over complementary subspaces. Compared to standard softmax attention~\citep{NIPS2017_3f5ee243}, which forms a \emph{single} attention map
$\mathrm{softmax}(\boldsymbol{Q}\boldsymbol{K}^{\intercal}/\sqrt{d_{k}})$ and applies it to $\boldsymbol{V}$, differential attention first splits queries/keys into two complementary channel subspaces, builds \emph{two} attention maps, and combines them via a learnable subtraction.

\paragraph{Input Linear Projections.}
Given tokenized features $\boldsymbol{X}\in\mathbb{R}^{N\times d_{\text{model}}}$, differential attention projects $\boldsymbol{X}$ into queries, keys, and values, and splits the query/key channels into two complementary subspaces:
\begin{equation}
    [\boldsymbol{Q}_{1};\!\boldsymbol{Q}_{2}]\!=\!\boldsymbol{X}\boldsymbol{W}^{Q}\!,\; [\boldsymbol{K}_{1};\!\boldsymbol{K}_{2}]\!=\!\boldsymbol{X}\boldsymbol{W}^{K}\!,\;
    \boldsymbol{V}\!=\!\boldsymbol{X}\boldsymbol{W}^{V}
    \label{eq:input_proj}
\end{equation}
where $\boldsymbol{W}^{Q}, \boldsymbol{W}^{K}, \boldsymbol{W}^{V}\in\mathbb{R}^{d_{\text{model}}\times d_k}$. Thus, $\boldsymbol{Q}_{1}$, $\boldsymbol{Q}_{2}$, $\boldsymbol{K}_{1}$, $\boldsymbol{K}_{2}\in\mathbb{R}^{N\times d_{k}/2}$ and $\boldsymbol{V}\in\mathbb{R}^{N\times d_{k}}$.

\paragraph{Differential Attention.}
Let $\lambda\in\mathbb{R}$ denote a learnable subtraction coefficient. Two softmax attention maps are computed as
\begin{equation}
    \boldsymbol{A}_{1} = \mathrm{softmax}(\frac{\boldsymbol{Q}_{1}\boldsymbol{K}_{1}^{\intercal}}{\sqrt{d_{k}/2}}) \quad \boldsymbol{A}_{2} = \mathrm{softmax}(\frac{\boldsymbol{Q}_{2}\boldsymbol{K}_{2}^{\intercal}}{\sqrt{d_{k}/2}}),
\end{equation}
and the output is formed by subtracting the second branch from the first:
\begin{equation}
    \mathrm{DiffAttn}(\boldsymbol{Q}, \boldsymbol{K}, \boldsymbol{V}, \lambda)
    = (\boldsymbol{A}_{1} - \lambda \boldsymbol{A}_{2})\boldsymbol{V}.
\end{equation}
Following~\citep{ye2025differential}, $\lambda$ is reparameterized as
\begin{equation}
    \lambda = \exp(\boldsymbol{\lambda}_{q_{1}}\boldsymbol{\lambda}_{k_{1}})
    - \exp(\boldsymbol{\lambda}_{q_{2}}\boldsymbol{\lambda}_{k_{2}})
    + \lambda_{\mathrm{init}},
\end{equation}
where $\boldsymbol{\lambda}_{q_{1}}, \boldsymbol{\lambda}_{k_{1}}, \boldsymbol{\lambda}_{q_{2}}, \boldsymbol{\lambda}_{k_{2}}\in\mathbb{R}^{d_k/2}$ are learnable vectors, and
\begin{equation}
    \lambda_{\mathrm{init}} = 0.8 - 0.6\exp(-0.3(l-1))
\end{equation}
is the layer-dependent initialization used in the original formulation.

\paragraph{Multi-Head Formulation.}
As in standard multi-head attention, differential attention is applied independently across $h$ heads. With per-head projection matrices
$\boldsymbol{W}^{Q}_{i}, \boldsymbol{W}^{K}_{i}, \boldsymbol{W}^{V}_{i}\in\mathbb{R}^{d_{\text{model}}\times d_h}$ and a layer-shared scalar $\lambda$, the $i$-th head is
\begin{equation}
    \mathrm{head}_{i}=
    \mathrm{DiffAttn}(\boldsymbol{Q}\boldsymbol{W}^{Q}_{i},
    \boldsymbol{K}\boldsymbol{W}^{K}_{i},
    \boldsymbol{V}\boldsymbol{W}^{V}_{i},
    \lambda).
\end{equation}
Following~\citep{ye2025differential}, the per-head output is normalized and rescaled as
\begin{equation}
    \mathrm{head}_{i}' = (1-\lambda_{\mathrm{init}})\,\mathrm{RMSNorm}(\mathrm{head}_{i}),
\end{equation}
and the multi-head output is obtained by concatenation:
\begin{equation}
    \mathrm{MultiHead}(\boldsymbol{Q}, \boldsymbol{K}, \boldsymbol{V}, \lambda)
    = [\mathrm{head}_{1}'; \dots; \mathrm{head}_{h}'].
\end{equation}

\paragraph{Final Output Projection.} 
The concatenated output is projected with $\boldsymbol{W}^{O}\in\mathbb{R}^{hd_{h}\times d_{\text{model}}}$.

\subsubsection{Kernelized (Linear) Attention}

Standard multi-head softmax attention has quadratic complexity $\mathcal{O}(N^{2})$ in the number of tokens $N$. Linear attention~\citep{pmlr-v119-katharopoulos20a} reduces this cost to $\mathcal{O}(N)$ by replacing softmax attention with a kernel feature map and exploiting associativity. Its input projections, multi-head structure, and output projection follow the standard attention formulation.

\paragraph{Linear Attention.}
Let $\phi:\mathbb{R}^{d_{k}}\rightarrow\mathbb{R}_{\ge 0}^{d_{k}}$ denote a nonnegative feature map. Linear attention replaces the softmax similarity with
\begin{equation}
    \langle \boldsymbol{Q}, \boldsymbol{K} \rangle_{\phi}
    = \phi(\boldsymbol{Q})\phi(\boldsymbol{K})^{\intercal}.
\end{equation}
The normalized output can be written as
\begin{equation}
    \mathrm{LinearAttn}(\boldsymbol{Q}, \boldsymbol{K}, \boldsymbol{V})
    =
    [\phi(\boldsymbol{Q})\phi(\boldsymbol{K})^{\intercal}]\,\boldsymbol{V}
    \oslash \boldsymbol{z},
\end{equation}
where $\oslash$ denotes element-wise division, and
\begin{equation}
    \boldsymbol{z}
    =
    \phi(\boldsymbol{Q})
    [\phi(\boldsymbol{K})^{\intercal}\mathbf{1}_{N}]
    \in \mathbb{R}^{N}.
\end{equation}
By associativity, the computation can be reordered as
\begin{equation}
    \mathrm{LinearAttn}(\boldsymbol{Q}, \boldsymbol{K}, \boldsymbol{V})
    =
    \phi(\boldsymbol{Q})
    [\phi(\boldsymbol{K})^{\intercal}\boldsymbol{V}]
    \oslash \boldsymbol{z},
\end{equation}
which yields linear complexity in $N$.

\subsection{Gated Differential Linear Attention Mixer}

We next describe the internal design of the GDLA mixer, which consists of three parts: a gated differential linear-attention path, a local token-mixing branch, and an output-fusion step that combines the two.

\subsubsection{Gated Differential Linear Attention}
\label{subsubsec:gdla}

\begin{figure}[th]
    \includegraphics[width=1.00\linewidth]{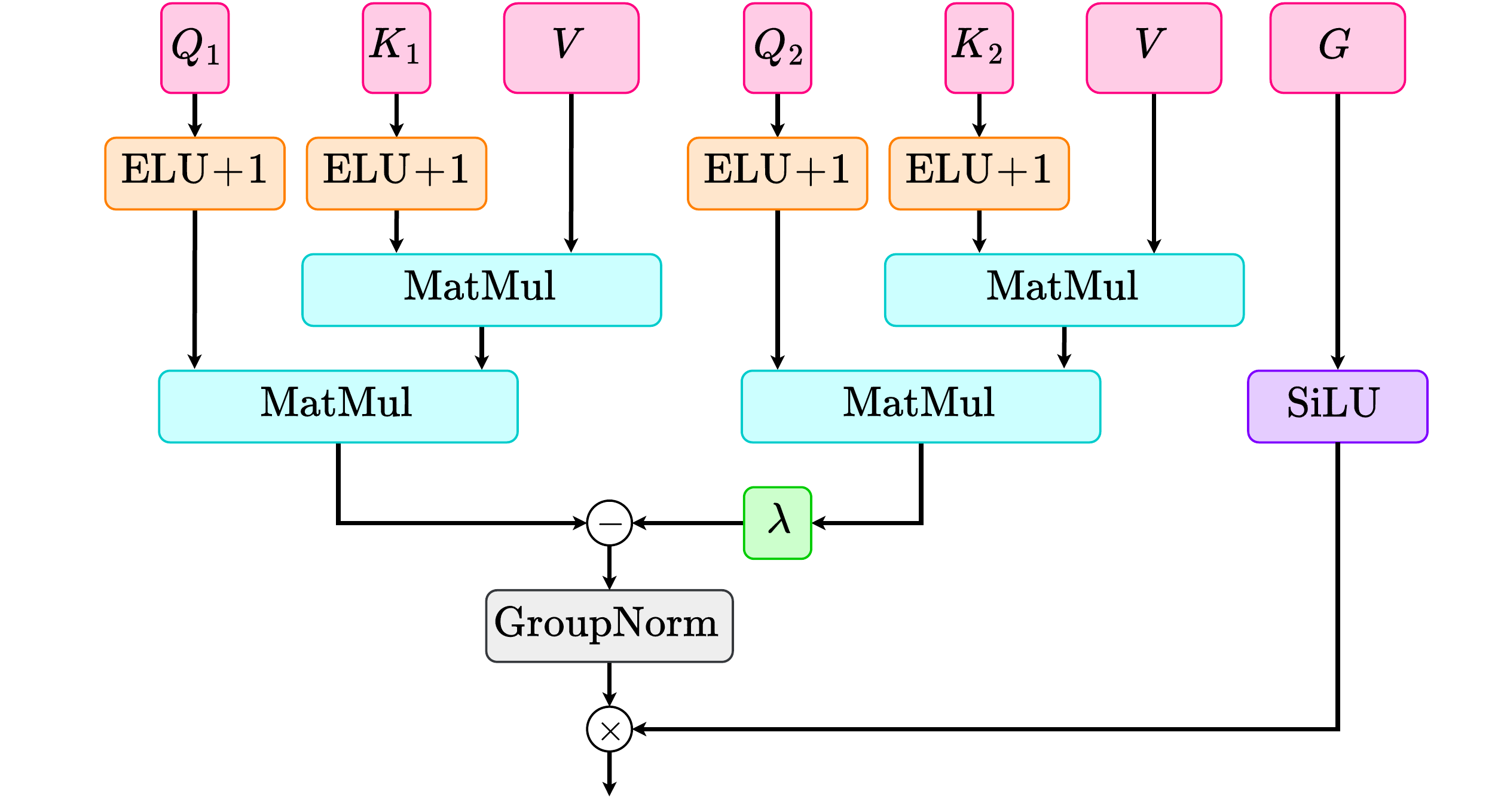}
    \vspace{-16pt}
    \caption{Gated Differential Linear Attention (GDLA). Two complementary query/key subspaces, $(\boldsymbol{Q}_{1},\boldsymbol{K}_{1})$ and $(\boldsymbol{Q}_{2},\boldsymbol{K}_{2})$, are derived from $\boldsymbol{X}$. Each branch applies kernelized linear attention with $\phi(\cdot)=\mathrm{ELU}(\cdot)+1$, and the resulting outputs are combined by learnable channel-wise subtraction, $\boldsymbol{A}_{1}-\boldsymbol{\lambda}\odot\boldsymbol{A}_{2}$, followed by \(\mathrm{RMSNorm}\) and a data-dependent SiLU gate.}
    \label{fig:gdla}
    \vspace{-10pt}
\end{figure}

Although linear attention efficiently captures long-range context, its nonnegative kernel features tend to produce diffuse, low-contrast aggregation, a phenomenon often referred to as \emph{attention dilution}~\citep{qin-etal-2022-devil}. In dense medical segmentation, such diffuse mixing can blur organ boundaries and weaken responses to thin or small structures. To address this issue, we adapt the subtractive principle of differential attention~\citep{ye2025differential} to the kernelized linear-attention setting. Unlike in softmax attention, the subtraction cannot be applied directly at the attention-map level because linear attention must retain its associative form to preserve linear complexity. We therefore compute two kernelized attention branches separately on complementary query/key subspaces and subtract their outputs after normalization. Figure~\ref{fig:gdla} illustrates this computation.

\paragraph{Differential Linear Attention.}
Using the input projections in Section~\ref{subsubse:diff-attn}, we obtain complementary query/key subspaces $\boldsymbol{Q}_{1}, \boldsymbol{Q}_{2}, \boldsymbol{K}_{1}, \boldsymbol{K}_{2}$ and values $\boldsymbol{V}$ (Eq.~\ref{eq:input_proj}). Using the feature map $\phi:\mathbb{R}^{d_{k/2}}\rightarrow\mathbb{R}_{\ge 0}^{d_{k/2}}$, we define

\begin{align}
    \boldsymbol{A}_{1}
    &= \phi(\boldsymbol{Q}_{1})
    [\phi(\boldsymbol{K}_{1})^{\intercal}\boldsymbol{V}]
    \oslash \boldsymbol{z}_{1}, \\
    \boldsymbol{A}_{2}
    &= \phi(\boldsymbol{Q}_{2})
    [\phi(\boldsymbol{K}_{2})^{\intercal}\boldsymbol{V}]
    \oslash \boldsymbol{z}_{2},
\end{align}
where
\begin{equation}
    \boldsymbol{z}_{i}
    =
    \phi(\boldsymbol{Q}_{i})
    [\phi(\boldsymbol{K}_{i})^{\intercal}\mathbf{1}_{N}],
    \quad i\in\{1,2\}.
\end{equation}
In our implementation, we use $\phi(\cdot)=\mathrm{ELU}(\cdot)+1$ to ensure nonnegativity. The two branch outputs are then combined through channel-wise subtraction:
\begin{equation}
    \mathrm{DiffLinearAttn}(\boldsymbol{Q}, \boldsymbol{K}, \boldsymbol{V}, \boldsymbol{\lambda})
    = \boldsymbol{A}_{1} - \boldsymbol{\lambda}\odot\boldsymbol{A}_{2},
\end{equation}
where $\boldsymbol{\lambda}\in\mathbb{R}^{d_{k}}$ is a learnable vector controlling the subtraction strength, and $\odot$ denotes the Hadamard product.

\paragraph{Multi-Head Differential Linear Attention.}
As in standard attention, we compute the above in parallel for \(h\) heads with per-head projections
\(\boldsymbol{W}^{Q}_{i},\boldsymbol{W}^{K}_{i},\boldsymbol{W}^{V}_{i}\in\mathbb{R}^{d_{\text{model}}\times d_h}\) and a per-head subtraction vector \(\boldsymbol{\lambda}_{i}\in\mathbb{R}^{d_{h}}\), $i\in\{1,\dots,h\}$. Let
\begin{equation}
\boldsymbol{Y}_{i}\!=\!\mathrm{DiffLinearAttn}(\boldsymbol{Q}\boldsymbol{W}^{Q}_{i}\!,\,\boldsymbol{K}\boldsymbol{W}^{K}_{i}\!,\,\boldsymbol{V}\mathbf{W}^{V}_{i}\!,\boldsymbol{\lambda}_{i}).
\end{equation}
We then stabilize gradients via RMS normalization
\begin{equation}
\hat{\boldsymbol{Y}}_{i}=\mathrm{RMSNorm}(\boldsymbol{Y}_{i}).
\end{equation}

\paragraph{Gating Mechanism.}
Kernelized attention remains a linear and relatively low-rank mapping, which can limit selectivity in dense prediction. We therefore apply a lightweight multiplicative gate to the per-head DLA output. For head $i$, we define
\begin{equation}
    \boldsymbol{G}_{i}=\mathrm{SiLU}(\boldsymbol{X}\boldsymbol{W}^{G}_{i})\in\mathbb{R}^{N\times d_{h}},
\end{equation}
where $\boldsymbol{W}^{G}_{i}\in\mathbb{R}^{d_{\text{model}}\times d_{h}}$ and $\mathrm{SiLU}(\cdot)$ denotes the SiLU activation. The gated head output is
\begin{equation}
    \mathrm{head}_{i}
    = \hat{\boldsymbol{Y}}_{i}\odot\boldsymbol{G}_{i}
    = \hat{\boldsymbol{Y}}_{i}\odot\mathrm{SiLU}(\boldsymbol{X}\boldsymbol{W}^{G}_{i}).
\end{equation}
We use a SiLU gate in the final model. As shown in Section~\ref{para:gate}, SiLU yielded slightly better empirical performance than sigmoid, so we adopt it as the default gating function.

\paragraph{Multi-Head Concatenation.}
Finally, we concatenate the head outputs along the channel dimension:
\begin{equation}
\mathrm{MultiHead}(\boldsymbol{Q},\boldsymbol{K},\boldsymbol{V},\boldsymbol{\lambda},\boldsymbol{G})\!=\![\mathrm{head}_{1};\dots ; \mathrm{head}_{h}].
\end{equation}

\subsubsection{Local Token Mixing Branch}
\label{subsubsec:token-mix}

While GDLA improves global selectivity, dense medical segmentation also depends heavily on short-range spatial continuity, especially around thin structures and object boundaries. To complement the global GDLA branch, we introduce a lightweight parallel \emph{local token-mixing} branch that explicitly aggregates neighborhood information. Following efficient dense-prediction designs~\citep{Cai_2023_ICCV}, we apply a depthwise convolution (DWC) to capture local spatial interactions and a pointwise convolution (PWC) to fuse channel information, yielding a local refinement path with modest overhead.

Let \(f\) denote the local mixer,
\begin{equation}
f(\cdot) = \mathrm{Conv}_{1\times 1}(\mathrm{DWC}_{3\times 3}(\cdot)),
\end{equation}
applied on the token grid (for 2D inputs, tokens are reshaped to $\mathcal{H}{\times}\mathcal{W}$ and then mapped back to length $N=\mathcal{H}\mathcal{W}$).

Given input $\boldsymbol{X}\in\mathbb{R}^{N\times d_{\text{model}}}$ and learned projections
$\boldsymbol{W}^{Q},\boldsymbol{W}^{K},\boldsymbol{W}^{V},\boldsymbol{W}^{G}\in\mathbb{R}^{d_{\text{model}}\times d_{k}}$,
we form the locally mixed inputs (splitting queries/keys into complementary subspaces as in GDLA):
\begin{align}
[\boldsymbol{Q}_{1}';\boldsymbol{Q}_{2}'\big] &= f(\boldsymbol{X}\boldsymbol{W}^{Q}) \quad
[\boldsymbol{K}_{1}';\boldsymbol{K}_{2}'\big] = f(\boldsymbol{X}\boldsymbol{W}^{K}) \\
\boldsymbol{V}' &= f(\boldsymbol{X}\boldsymbol{W}^{V}) \quad
\boldsymbol{G}' = f(\boldsymbol{X}\boldsymbol{W}^{G}),
\end{align}
where $\boldsymbol{Q}_{1}',\boldsymbol{Q}_{2}',\boldsymbol{K}_{1}',\boldsymbol{K}_{2}'\in\mathbb{R}^{N\times d_{k}/2}$ and $\boldsymbol{V}',\boldsymbol{G}'\in\mathbb{R}^{N\times d_{k}}$.
The subsequent multi-head computation and gating follow GDLA verbatim, substituting the primed tensors for their unprimed counterparts.

\begin{table*}[th]
\caption{Evaluation results on the Synapse dataset (\textcolor{blue}{blue} indicates the best and \textcolor{red}{red} the second best results). Results are averaged over three runs.}
\begin{center}
\vspace{-18pt}
\resizebox{0.9\textwidth}{!}{
\begin{tabular}{l|r|r|cccccccc|cc}
    \toprule
    \multirow{2}{*}{Methods}& \multirow{2}{*}{Params}& \multirow{2}{*}{FLOPs}& \multirow{2}{*}{Spl.}&  \multirow{2}{*}{RKid.}& \multirow{2}{*}{LKid.}&  \multirow{2}{*}{Gal.}&  \multirow{2}{*}{Liv.}&  \multirow{2}{*}{Sto.}& \multirow{2}{*}{Aor.}& \multirow{2}{*}{Pan.}& \multicolumn{2}{c}{Average}\\
    \cmidrule{12-13}
    & & & & & & & & & & &DSC$\uparrow$ &HD95$\downarrow$\\

    \midrule
    R50+U-Net ~\cite{chen2021transunet}& 30.42 M& -&85.87 &78.19 &80.60 &63.66 &93.74 &74.16 &87.74  &56.90 &74.68 &36.87\\


    TransUNet~\cite{chen2021transunet}& 96.07 M& 88.91 G&85.08 &77.02 &81.87 &63.16 &94.08 &75.62 &87.23  &55.86 &77.49 &31.69\\


    Swin-UNet~\cite{cao2021swinunet}& 27.17 M& 6.16 G& 90.66& 79.61& 83.28& 66.53& 94.29& 76.60& 85.47& 56.58& 79.13& 21.55\\



    HiFormer-B~\cite{heidari2023hiformer}& 25.51 M& 8.05 G& 90.99& 79.77& 85.23& 65.23& 94.61& 81.08& 86.21& 59.52& 80.39& 14.70\\



    VM-UNet~\cite{ruan2024vm}& 44.27 M& 6.52 G&  89.51&  82.76& 86.16& 69.41& 94.17& 81.40& 86.40&  58.80& 81.08& 19.21\\



    PVT-EMCAD-B2~\cite{rahman2024emcad} & 26.76 M & 5.60 G & 92.17& 84.10& 88.08& 68.87& 95.26& 83.92& 88.14 & 68.51& 83.63& 15.68\\

    MSA$^{\text{2}}$Net~\cite{kolahi2024msa} & 112.77 M& 15.56 G& 92.69& 84.24& 88.30& \textcolor{blue}{74.35}& 95.59& 84.03& \textcolor{red}{89.47}& 69.30& 84.75& 13.29\\


    2D D-LKA Net~\cite{azad2023selfattention} & 101.64 M & 19.92 G & 91.22& 84.92& 88.38& \textcolor{red}{73.79}& 94.88& 84.94& 88.34& 67.71& 84.27& 20.04\\

    CENet~\citep{BozAfs_CENet_MICCAI2025} & 33.39 M & 12.76 G & \textcolor{blue}{93.58}& \textcolor{red}{85.08}& \textcolor{blue}{91.18}& 68.29& 95.92& 81.68& 89.19& \textcolor{red}{70.71}& \textcolor{red}{85.04}& \textcolor{blue}{8.84}\\

    PVT-DiffAttn~\citep{ye2025differential} & 28.75 M & 5.61 G & 92.17& 83.43& 86.94& 68.64& 95.73& \textcolor{red}{85.74}& 88.28& 68.94& 83.73& 23.24 \\

    PVT-SelfAttn~\citep{NIPS2017_3f5ee243} & 28.25 M & 5.40 G & 91.72& \textcolor{blue}{85.67}& 87.20& 67.88& 95.92& 85.00& \textcolor{blue}{89.79}& 66.93& 83.77& 22.40 \\

    PVT-LinearAttn~\citep{pmlr-v119-katharopoulos20a} & 28.25 M & 5.38 G & 90.46 & 83.22 & 87.02 & 69.00 & \textcolor{red}{95.92} & 85.54 & 88.23 & 67.28 & 83.33 & 18.62 \\

    \midrule
    \textbf{\name} & 32.13 M & 6.85 G &\textcolor{red}{92.96} & 85.03 & \textcolor{red}{90.16} & 69.97 & \textcolor{blue}{96.32} & \textcolor{blue}{88.99} & 88.26 & \textcolor{blue}{70.86} & \textcolor{blue}{85.32} & \textcolor{red}{12.41}\\



    \bottomrule
\end{tabular}}
\vspace{-10pt}
\end{center}    
\label{tab:synapse}
\end{table*}

\begin{figure*}[th]
    \centering
    \includegraphics[width=\linewidth]{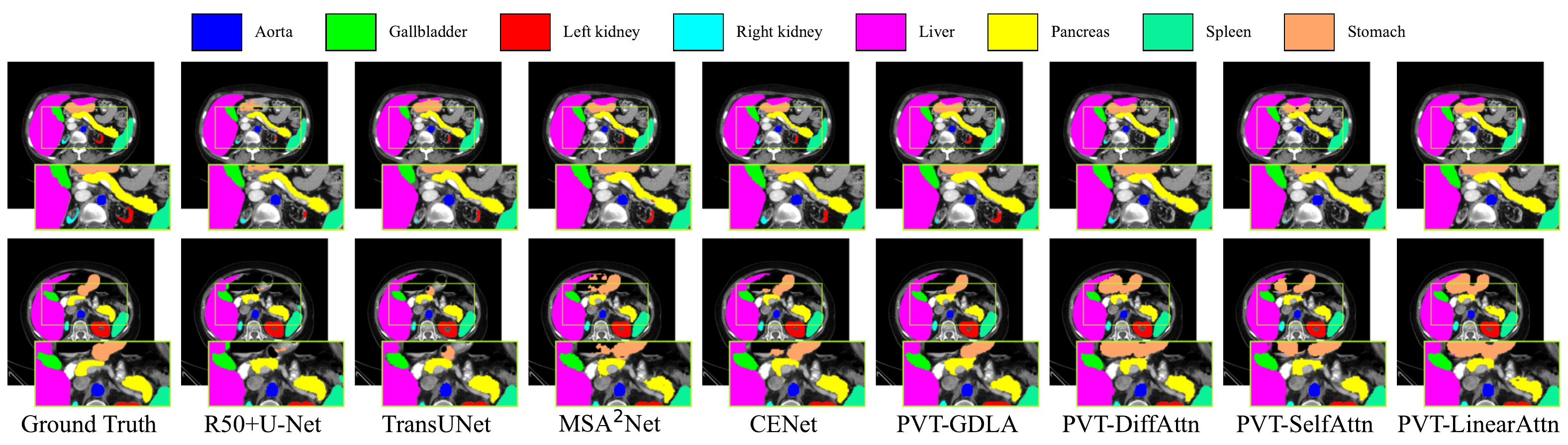}
    \vspace{-18pt}
    \caption{Qualitative results of the proposed method versus other approaches on the {\it Synapse} dataset.}
    \label{fig:synapse}
    \vspace{-5pt}
\end{figure*}

\subsubsection{Output Fusion}
We fuse the outputs of the global GDLA branch and the local token-mixing branch through a final linear projection. Let
\begin{align}
    \hat{\boldsymbol{Y}} &= \mathrm{MultiHead}(\boldsymbol{Q}, \boldsymbol{K}, \boldsymbol{V}, \boldsymbol{\lambda}, \boldsymbol{G}), \\
    \hat{\boldsymbol{Y}}' &= \mathrm{MultiHead}(\boldsymbol{Q}', \boldsymbol{K}', \boldsymbol{V}', \boldsymbol{\lambda}', \boldsymbol{G}'),
\end{align}
denote the global and local branch outputs, respectively. We concatenate them along the channel dimension and apply a linear projection:
\begin{equation}
    \boldsymbol{O}
    =
    \mathrm{Concat}(\hat{\boldsymbol{Y}}, \hat{\boldsymbol{Y}}')\boldsymbol{W}^{O}
    \in \mathbb{R}^{N\times d_{\text{model}}},
\end{equation}
where $\boldsymbol{W}^{O}\in\mathbb{R}^{2d_{\text{model}}\times d_{\text{model}}}$. This fusion preserves linear complexity while combining globally selective context modeling with explicit local refinement.

\subsection{Feed-Forward Network}
\label{subsec:ffn}

We evaluate three feed-forward network (FFN) variants within the decoder block: (i) a standard two-layer MLP with $4\times$ expansion~\citep{NIPS2017_3f5ee243}, (ii) $\mathrm{SwiGLU}$~\citep{shazeer2020glu}, and (iii) a Mix-FFN variant that introduces depthwise convolution for additional local token interaction~\citep{xie2025sana}. In our experiments, Mix-FFN provided the best trade-off between accuracy and efficiency, likely because the depthwise convolution complements GDLA by injecting additional neighborhood-level mixing.

Let $\boldsymbol{X}\in\mathbb{R}^{N\times d_{\text{model}}}$ and $d_{\text{hidden}}=\alpha d_{\text{model}}$ with $\alpha=4$ unless otherwise noted. The Mix-FFN is defined as:
\begin{align}
    [\hat{\boldsymbol{X}}; \boldsymbol{G}]
    &= \mathrm{DWC}_{3\times3}\!\left(\mathrm{SiLU}(\mathrm{Conv}_{1\times1}(\boldsymbol{X}))\right)
    \in\mathbb{R}^{N\times 2d_{\text{hidden}}}, \nonumber \\
    \boldsymbol{Y}
    &= \mathrm{Conv}_{1\times1}(\hat{\boldsymbol{X}}\odot\mathrm{SiLU}(\boldsymbol{G}))
    \in\mathbb{R}^{N\times d_{\text{model}}},
\end{align}
where $\odot$ denotes element-wise multiplication. We use $k=3$ in the main experiments; larger kernels are possible at modest additional cost.
\section{Experiment}
\label{sec:exp}

We evaluate \name on five 2D medical image segmentation benchmarks spanning multiple imaging modalities, including CT, MRI, ultrasound, and dermoscopy. Unless otherwise stated, our implementation is based on PyTorch and trained on a single NVIDIA L40S GPU (46\,GB) using an input resolution of $224\times224$ and the BDoU loss~\citep{rahman2023multi}. We report quantitative results across all datasets and complement them with qualitative visualizations and controlled ablation studies. Detailed dataset splits, optimization settings, and training schedules are provided in the Supplementary Section~\ref{appsec:exp}.

\begin{table*}[ht]
\centering
\captionof{table}{Evaluation results on the \textit{ACDC} (left), \textit{BUSI} (middle), and skin–lesion benchmarks \textit{PH$^2$} and \textit{HAM10000} (right). All metrics are averaged over three runs.}
\vspace{-5pt}
\resizebox{\textwidth}{!}{
\begin{tabular}{l||c||r}
    \begin{tabular}{lcccc}  
        \toprule
        \textbf{Methods} & \textbf{Avg. Dice} & \textbf{RV} & \textbf{MYO} & \textbf{LV} \\
        \midrule
        R50+UNet~\citep{chen2021transunet} & 87.55 & 87.10 & 80.63 & 94.92\\
        R50+AttnUNet~\citep{chen2021transunet} & 86.75 & 87.58 & 79.20 & 93.47\\
        TransUNet~\citep{chen2021transunet}& 89.71 & 88.86 &  84.53 & 95.73 \\
        Swin-UNet~\citep{cao2021swinunet} & 90.00 & 88.55 & 85.62 & 95.83 \\
        R50+ViT+CUP~\citep{chen2021transunet} & 87.57 & 86.07 & 81.88 & 94.75 \\
        MT-UNet~\citep{wang2022mixed} & 90.43 & 86.64 & 89.04 & 95.62 \\
        MISSFormer~\citep{huang2022missformer} & 90.86 & 89.55 & 88.04 & 94.99 \\
        PVT-EMCAD-B2~\citep{rahman2024emcad} & 92.12 & 90.65 & 89.68 & 96.02 \\
        CENet~\citep{BozAfs_CENet_MICCAI2025} & 92.18 & \textcolor{red}{90.90} & 89.63 & 95.99 \\
        PVT-DiffAttn~\citep{ye2025differential} & \textcolor{red}{92.20} & 90.64 & \textcolor{red}{89.90} & 96.06 \\
        PVT-SelfAttn~\citep{NIPS2017_3f5ee243} & 92.10 & 90.47 & 89.76 & \textcolor{red}{96.06} \\
        PVT-LinearAttn~\citep{pmlr-v119-katharopoulos20a} & 91.77 & 89.80 & 89.52 & 96.00 \\
        \midrule
        \textbf{\name} & \textcolor{blue}{92.53} & \textcolor{blue}{91.30} & \textcolor{blue}{90.21} & \textcolor{blue}{96.07} \\
        \bottomrule  
    \end{tabular}
    \label{tab:acdc}
    &
    \begin{tabular}{lc}  
        \toprule
        \textbf{Methods} & \textbf{Avg. Dice} \\
        \midrule
        U-Net~\citep{ronneberger2015u} & 74.04 \\
        PraNet~\citep{fan2020pranet} & 75.41 \\
        CaraNet~\citep{lou2022caranet} & 77.34 \\
        AttnUNet~\citep{oktay2018attention} & 74.48 \\
        DeepLabv3+~\citep{azad2020attention} & 76.81 \\
        TransUNet~\citep{chen2021transunet}& 78.30 \\
        TransFuse~\citep{zhang2021transfuse} & 79.36 \\
        Swin-UNet~\citep{cao2021swinunet} & 77.38 \\
        PVT-EMCAD-B2~\citep{rahman2024emcad} & \textcolor{red}{80.25} \\
        PVT-DiffAttn~\citep{ye2025differential} & 79.84 \\
        PVT-SelfAttn~\citep{NIPS2017_3f5ee243} & 79.42 \\
        PVT-LinearAttn~\citep{pmlr-v119-katharopoulos20a} & 79.14 \\
        \midrule
        \textbf{\name} & \textcolor{blue}{80.54} \\
        \bottomrule  
    \end{tabular}
    \label{tab:busi}
    &
    \resizebox{0.963\columnwidth}{!}{%
    \begin{tabular}{lcccc}  
        \toprule
        \multirow{2}{*}{\textbf{Methods}}& \multicolumn{2}{c}{\textbf{PH$^2$}} & \multicolumn{2}{c}{\textbf{HAM10000}} \\
        \cmidrule{2-5}
        & \textbf{Dice} & \textbf{Acc.} & \textbf{Dice} & \textbf{Acc.} \\
        \cmidrule{1-5}
        U-Net~\citep{ronneberger2015u} & 89.36 & 92.33 & 91.67 & 95.67 \\
        TransUNet~\citep{chen2021transunet} & 88.40 & 92.00 & 93.53 & 96.49 \\
        Swin-Unet~\citep{cao2021swinunet} & 94.49 & 96.78 & 92.63 & 96.16 \\
        DeepLabv3+~\citep{chen2018encoder} & 92.02 & 95.03 & 92.51 & 96.07 \\
        Att-UNet~\citep{oktay2018attention} & 90.03 & 92.76 & 92.68 & 96.10 \\
        UCTransNet~\citep{wang2022uctransnet} & 90.93 & 94.08 & 93.46 & 96.84 \\
        MissFormer~\citep{huang2022missformer} & 85.50 & 90.50 & 92.11 & 96.21 \\
        CENet~\citep{BozAfs_CENet_MICCAI2025} & 95.04 & \textcolor{red}{97.19} & 94.71 & 97.04 \\
        PVT-DiffAttn~\citep{ye2025differential} & \textcolor{red}{95.10} & 97.13 & 94.73 & \textcolor{red}{97.10} \\
        PVT-SelfAttn~\citep{NIPS2017_3f5ee243} & 95.08 & 97.18 & \textcolor{red}{94.78} & 97.06 \\
        PVT-LinearAttn~\citep{pmlr-v119-katharopoulos20a} & 94.78 & 96.68 & 94.68 & 96.87\\
        \midrule
        \textbf{\name} & \textcolor{blue}{95.59} & \textcolor{blue}{97.20} & \textcolor{blue}{95.01} & \textcolor{blue}{97.38} \\
        \bottomrule
    \end{tabular}}
    \label{tab:ph2-ham10k}
    \\
\end{tabular}}
\label{tab:acdc-busi-skin}
\end{table*}

\subsection{Results}

\paragraph{Radiology.} The performance of \name on radiological datasets is evaluated, with Table~\ref{tab:synapse} presenting the quantitative results on the Synapse CT dataset using Dice score and HD metrics, and Table~\ref{tab:acdc} showing the quantitative results on the ACDC dataset~\citep{bernard2018deep}. Our approach outperforms existing CNN-based, Transformer-based, and hybrid models. \name achieves the SoTA performance with comparable number of parameters but lower FLOPs. These results underscore \name's ability in segmenting various organs. Figure~\ref{fig:synapse} provides a visual representation of \name's performance in segmenting various organs, demonstrating \name's accuracy in multi-scale segmentation of the liver, stomach, pancreas, spleen, and kidneys. Additionally, \name also shows effectiveness on the ACDC dataset for cardiac segmentation in the MRI image, achieves the highest average Dice score of $92.53\%$, and excels in all three organ segmentation tasks.

\paragraph{Ultrasound.}
We evaluate on the BUSI breast ultrasound dataset~\citep{al2020dataset}. As summarized in Table~\ref{tab:busi}, \name attains the best average Dice of 80.54\% surpassing prior PVT-based baseline (PVT-EMCAD-B2) by +0.29\% and outperforms other PVT decoders, PVT-DiffAttn, PVT-SelfAttn, and PVT-LinearAttn, by +0.70\%, +1.12\%, and +1.40\% Dice, respectively.

\begin{figure}[th]
    \centering
    \includegraphics[width=\linewidth]{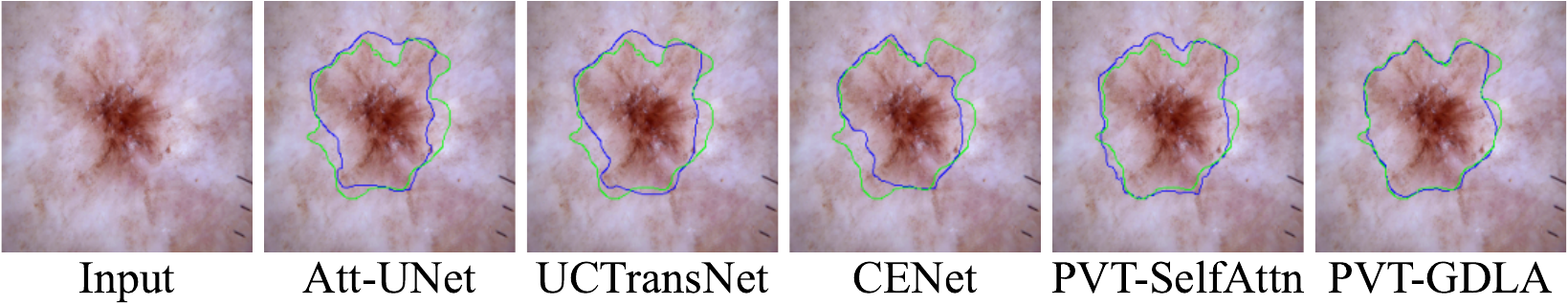}
    \\ \vspace{-0.775em}
    \includegraphics[width=\linewidth]{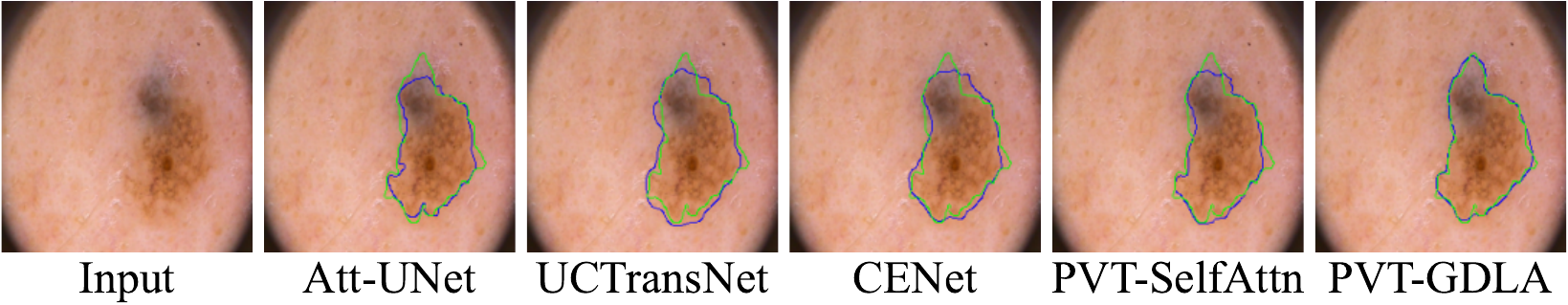}
    \vspace{-15pt}
    \caption{Visual comparison of the proposed method versus others on the Skin datasets. \textit{HAM10000} (top) and \textit{PH$^{2}$} (bottom). The blue and the green lines represent prediction and ground truth respectively.}
    \label{fig:skin}
\end{figure}

\paragraph{Dermoscopy.}
Table~\ref{tab:acdc-busi-skin} evaluates the proposed network on two skin lesion segmentation datasets, {\it HAM10000}~\citep{tschandl2018ham10000} and {\it PH$^{2}$}~\citep{mendoncca2015ph2} using Dice and accuracy metrics. \name outperforms CNN-based, Transformer-based, and hybrid methods, showing strong performance and generalization across both dermoscopy benchmarks. Moreover, Figure~\ref{fig:skin} showcases \name's capability in capturing intricate structures and producing precise boundaries through effective boundary integration.

\subsection{Attention Visualization}


\begin{figure}[ht]
    \centering
    \begin{tabular}{cc}
        \begin{minipage}{0.19\linewidth}
            \centering
            {\fontsize{6}{7}\selectfont Input}
            \includegraphics[width=\linewidth]{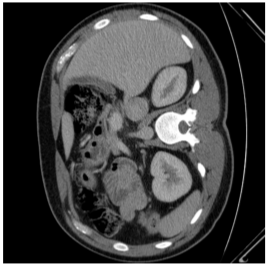}
            \vspace{-1.6em}
            \\
            {\fontsize{6}{7}\selectfont Ground Truth}
            \includegraphics[width=\linewidth]{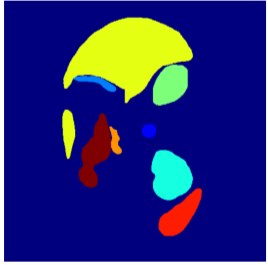}
        \end{minipage}
        & \hspace{-1.2em}
        \begin{minipage}{0.78\linewidth}
            \centering
            \includegraphics[width=\linewidth]{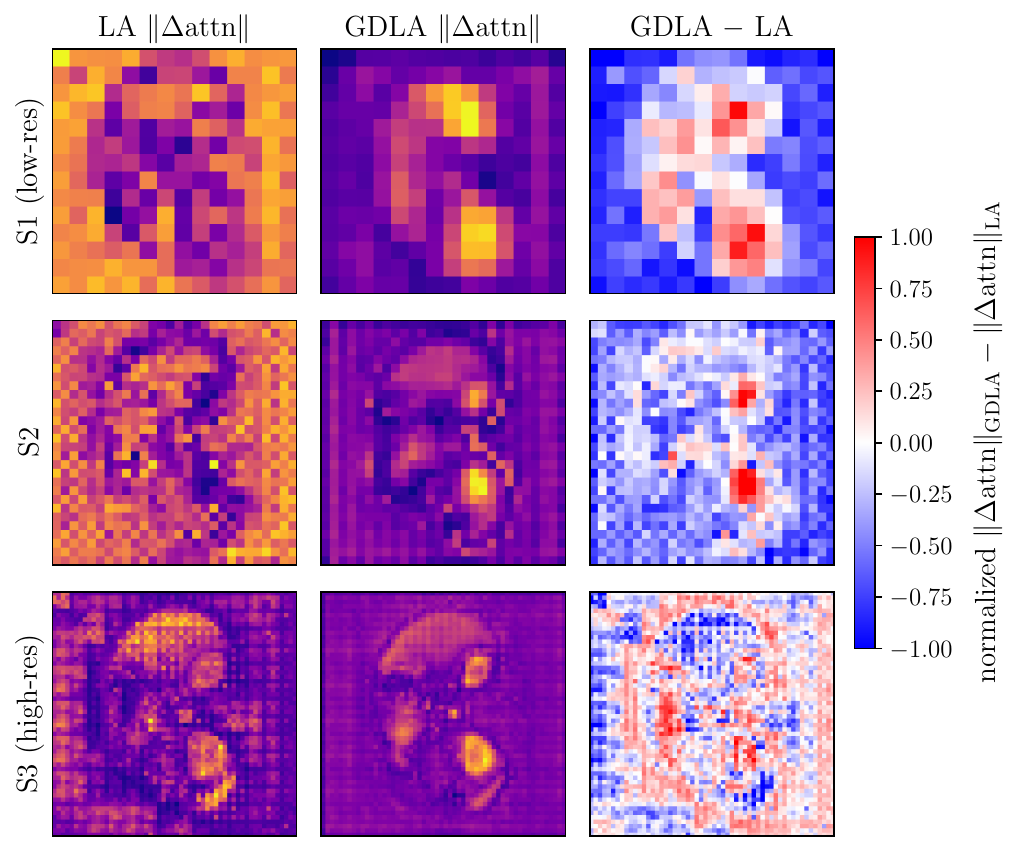}
        \end{minipage}
    \end{tabular}
    \vspace{-5pt}
    \caption{Linear Attention (LA) vs. Gated Differential Linear Attention (GDLA) on a \textit{Synapse} CT slice. Rows correspond to decoder stages (coarse $\rightarrow$ fine); columns show token-norm maps $\|\text{input}\|$ and update magnitudes $\|\Delta\text{attn}\|$ for LA (left) and GDLA (right), with a difference panel visualizing normalized $\|\Delta\text{attn}\|$ (GDLA$-$LA).}
    \label{fig:delta-attn}
\end{figure}

Across decoder stages, Figure~\ref{fig:delta-attn} qualitatively compares linear attention (LA) and gated differential linear attention (GDLA) mixer on a Synapse CT slice. LA exhibits relatively diffuse token activations and noisy update maps, whereas GDLA mixer produces sharper and more anatomically structured responses that better align with organ regions. In the LA columns, the input norm maps $\|\text{input}\|$ appear spatially smeared, and the corresponding $\|\Delta\text{attn}\|$ maps contain substantial high-frequency speckle, suggesting attention updates spread broadly across the field of view. By contrast, GDLA mixer yields clearer organ silhouettes across decoder levels, and its $\|\Delta\text{attn}\|$ maps are more concentrated around organ interiors and boundaries, which is qualitatively consistent with reduced attention dilution.


\begin{figure}[ht]
    \centering
    \begin{tabular}{cc}
        \begin{minipage}{0.19\linewidth}
            \centering
            {\fontsize{6}{7}\selectfont Input}
            \includegraphics[width=\linewidth]{sec/4_experiment/imgs/input.png}
            \vspace{-1.6em}
            \\
            {\fontsize{6}{7}\selectfont Ground Truth}
            \includegraphics[width=\linewidth]{sec/4_experiment/imgs/GT.png}
        \end{minipage}
        & \hspace{-1.2em}
        \begin{minipage}{0.78\linewidth}
            \centering
            \includegraphics[width=\linewidth]{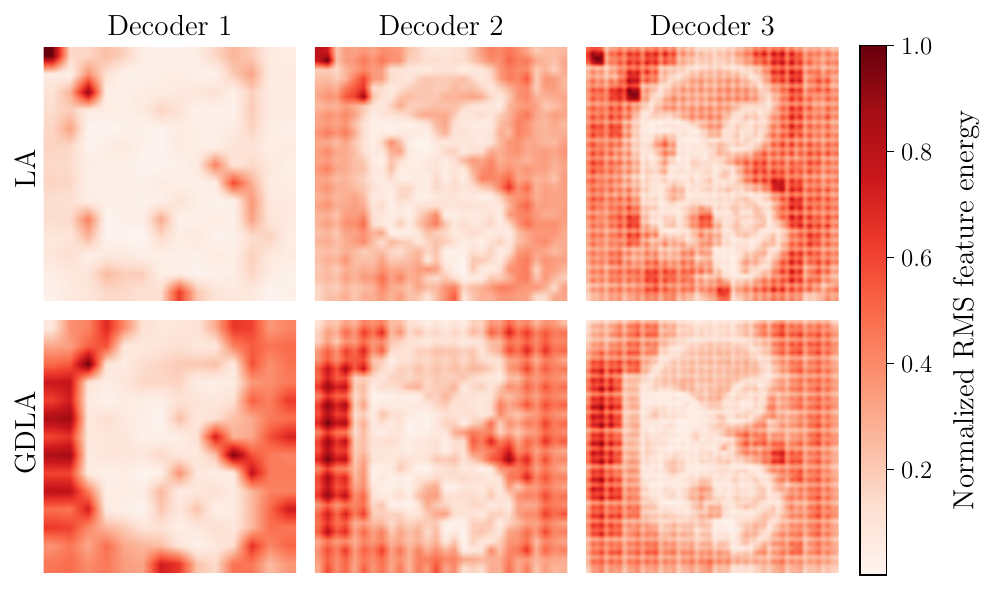}
        \end{minipage}
    \end{tabular}
    \vspace{-5pt}
    \caption{Channel saliency across decoder stages. Normalized RMS feature--energy per channel for Decoder 1--3 (coarse to fine) comparing linear attention (LA) and GDLA. The left column includes the input data and ground truth segmentation.}
    \label{fig:csm}
\end{figure}

Figure~\ref{fig:csm} shows the Channel Saliency Map (CSM), which measures the root-mean-square (RMS) feature energy per token at each decoder stage. Under LA, the first token tends to exhibit disproportionately high energy relative to other positions, a pattern consistent with the attention-sink phenomenon. In addition, the contrast between foreground and background regions is relatively weak, reflecting the diffuse behavior seen in Figure~\ref{fig:delta-attn}. In comparison, GDLA mixer yields a more balanced and structure-aware energy distribution and avoids excessive concentration on a small number of tokens, especially the first one. These observations corroborate the qualitative trends in Figure~\ref{fig:delta-attn}.

\subsection{Ablation Study}

\subsubsection{Effect of Different Components}

\begin{table}[h]
\small
\begin{center}
    \caption{Effect of different components of \name on the \textit{PH}$^{2}$ datasets. All results are averaged over three runs. The best result is shown in bold.}
    \vspace{-8pt}
    \resizebox{\columnwidth}{!}{%
    \begin{tabular}{ccccccc}  
        \toprule
        \multicolumn{3}{c}{\textbf{Components}} & \multirow{2}{*}{\textbf{Params}} & \multirow{2}{*}{\textbf{FLOPs}} & \multirow{2}{*}{\textbf{Dice}} \\
        DiffAttn & Mixer & Gate & & & \\
        \midrule
        \xmark & \xmark & \xmark & 28.25 M & 5.38 G & 94.78 \\
        \cmark & \xmark & \xmark & 29.89 M & 6.05 G & 95.08 \\
        \cmark & \xmark & \cmark & 30.01 M & 6.10 G & 95.20 \\
        \cmark & \cmark & \xmark & 31.13 M & 6.52 G & 95.46 \\
        \cmark & \cmark & \cmark & 32.13 M & 6.85 G & \textbf{95.59} \\
        \bottomrule
    \end{tabular}}
    \label{tab:ph2-ablation}
\end{center}
\end{table}

We assess the contribution of GDLA's components on \textit{PH}$^{2}$ through controlled ablations against a linear-attention baseline. As summarized in Table~\ref{tab:ph2-ablation}, both $\mathrm{DiffAttn}$ and the local token-mixing branch provide the largest gains: adding $\mathrm{DiffAttn}$ improves Dice by approximately $0.3\%$, while introducing the mixer yields a further gain of about $0.4\%$. Notably, $\mathrm{DiffAttn}$ introduces less parameter and FLOPs overhead than the local mixer. Combined, the two components improve Dice by roughly $0.7\%$ over the baseline. The $\mathrm{SiLU}$ gate yields a smaller but still positive gain. Overall, the full GDLA configuration achieves the best result, improving Dice by approximately $0.8\%$ over the linear-attention baseline.

\subsubsection{Architectural Choices}

We further examine several design choices within the proposed GDLA mixer-based decoder, including kernel sizes in the local refinement path, the gating function used in the GDLA branch, and the FFN design within the decoder block. The following ablations evaluate how these choices affect segmentation performance on \textit{Synapse}, \textit{ACDC}, and \textit{HAM10000}.

\begin{table}[!ht]
\small
\begin{center}
\caption{Effect of $\mathrm{ConvTranspose}$ kernel size in Upsample and depth-wise convolution kernel size in local token mixing on \textit{Synapse} multi-organ, \textit{ACDC}, \textit{HAM10000} datasets. All results are averaged over three runs. Best results are highlighted in bold.}
\resizebox{\columnwidth}{!}{%
\begin{tabular}{cc|cccc}  
    \toprule
    \multicolumn{2}{c}{\textbf{Kernel Size}} & \multicolumn{3}{c}{\textbf{Avg. Dice}} \\
    \midrule
    $\mathrm{DWC}$ & $\mathrm{ConvTranspose}$ & \textbf{Synapse} & \textbf{ACDC} & \textbf{HAM1000} \\
    \midrule
    $3{\times}3$ & $3{\times}3$ & \textbf{85.32} & \textbf{92.53} & \textbf{95.01} \\ 
    $3{\times}3$ & $5{\times}5$ & 84.94 & 92.23 & 94.73 \\
    $5{\times5}$ & $3{\times}3$ & 84.73 & 92.21 & 94.68 \\
    $5{\times5}$ & $5{\times}5$ & 84.63 & 92.17 & 94.62 \\
    \bottomrule
\end{tabular}
\label{tab:kernel}
}
\end{center}
\end{table}

\paragraph{Kernel Sizes.}
As shown in Table~\ref{tab:kernel}, we jointly evaluate the kernel sizes of the local token mixer ($\mathrm{DWC}$) and the upsampling layer ($\mathrm{ConvTranspose}$). The $3{\times}3$ / $3{\times}3$ configuration achieves the best average Dice across all three datasets. Increasing either kernel to $5{\times}5$ leads to consistent but modest performance drops. We therefore adopt $3{\times}3$ kernels for both components.

\begin{table}[ht]
\small
\centering
\caption{Effect of gating functions in GDLA. Comparison of sigmoid and SiLU gating, evaluated by average Dice (\%, mean$\pm$std) on \textit{Synapse}, \textit{ACDC}, and \textit{HAM10000}.}
\vspace{-5pt}
\begin{tabular}{cccc}
    \toprule
    \multirow{2}{*}{\textbf{Gate}} & \multicolumn{3}{c}{\textbf{Avg. Dice}} \\
    \cmidrule{2-4}
    & \textbf{Synapse} & \textbf{ACDC} & \textbf{HAM10000} \\
    \midrule
    Sigmoid & 84.94 & 92.24 & 94.70 \\
    SiLU    & \textbf{85.32} & \textbf{92.53} & \textbf{95.01} \\
    \bottomrule
\end{tabular}
\vspace{-10pt}
\label{tab:gate}
\end{table}

\paragraph{Gating Function.}
\label{para:gate}
We compare different gating functions for the GDLA module in Table~\ref{tab:gate} to assess whether gate selection materially affects performance. SiLU consistently outperforms sigmoid across all three datasets, which supports its use in the final model.

\begin{table}[ht]
\small
\begin{center}
\caption{Effect of FFN design within GDLA. Comparison of three feed-forward variants, 2-layer MLP ($4\times$ expansion), SwiGLU, and Mix-FFN, evaluated by average Dice (\%) on \textit{Synapse}, \textit{ACDC}, and \textit{HAM10000}.}
\vspace{-5pt}
\begin{tabular}{cccc}  
    \toprule
    \multirow{2}{*}{\textbf{FFN}} & \multicolumn{3}{c}{\textbf{Avg. Dice}} \\
    \cmidrule{2-4}
    & \textbf{Synapse} & \textbf{ACDC} & \textbf{HAM10000} \\
    \midrule
    2-layer MLP & 84.73 & 92.33 & 94.84 \\
    SwiGLU & 84.78 & 92.25 & 94.90 \\
    Mix-FFN & \textbf{85.32} & \textbf{92.53} & \textbf{95.01} \\
    \bottomrule
\end{tabular}
\vspace{-20pt}
\label{tab:ffn}
\end{center}
\end{table}

\paragraph{Feed-Forward Network.}
Finally, we evaluate the three FFN variants introduced in Section~\ref{subsec:ffn}, with results reported in Table~\ref{tab:ffn}. Mix-FFN performs best among the evaluated variants, suggesting that the additional local token interaction introduced by depthwise convolution complements GDLA and benefits boundary-sensitive dense prediction.
\section{Conclusion}

We introduced \emph{Gated Differential Linear Attention} (GDLA) mixer, a linear-attention module for medical image segmentation. GDLA mixer combines a gated differential linear-attention path for globally selective feature aggregation with a parallel local token-mixing path for short-range refinement, and fuses the two while retaining linear-time complexity. Across multiple 2D medical segmentation benchmarks, models instantiated with GDLA mixer achieve consistent improvements and a favorable accuracy--efficiency trade-off.

{
    \small
    \bibliographystyle{ieeenat_fullname}
    \bibliography{main}
}


\clearpage
\setcounter{page}{1}
\maketitlesupplementary

\section{Preliminaries}

\subsection{Multi-Head Softmax Attention}

Let the input be $\boldsymbol{X}\in\mathbb{R}^{N\times d_{\text{model}}}$, where $N$ is the number of tokens (e.g., $N=\mathcal{H}\mathcal{W}$ for a 2D grid) and $d_{\text{model}}$ is the channel width. The softmax attention~\citep{NIPS2017_3f5ee243} operates with the following steps.

\paragraph{Input Linear Projections.}
The input $\boldsymbol{X}$ is linearly transformed into queries $\boldsymbol{Q}$, keys $\boldsymbol{K}$, and values $\boldsymbol{V}\in\mathbb{R}^{N\times d_{k}}$ via learned weight matrices $\boldsymbol{W}^Q,\boldsymbol{W}^K,\boldsymbol{W}^V\in\mathbb{R}^{d_{\text{model}}\times d_{k}}$,
\begin{equation}
\boldsymbol{Q}=\boldsymbol{X}\boldsymbol{W}^{Q},\quad
\boldsymbol{K}=\boldsymbol{X}\boldsymbol{W}^{K},\quad
\boldsymbol{V}=\boldsymbol{X}\boldsymbol{W}^{V}.
\end{equation}

\paragraph{Scaled Dot-Product (Softmax) Attention.}
The scaled dot-product attention (SDPA) is computed as
\begin{equation}
    \mathrm{Attention}(\boldsymbol{Q}, \boldsymbol{K}, \boldsymbol{V}) = \mathrm{softmax}\left(\frac{\boldsymbol{Q}\boldsymbol{K}^{\intercal}}{\sqrt{d_{k}}}\right)\boldsymbol{V}\in\mathbb{R}^{N\times d_{k}},
\end{equation}
where $\boldsymbol{Q}\boldsymbol{K}^{\intercal}/\sqrt{d_{k}}$ is the similarity matrix and $\mathrm{softmax}(\cdot)$ normalizes each row into a probability distribution.

\paragraph{Multi-Head Concatenation.}
In multi-head attention, this computation is performed in parallel across $h$ heads. Each head uses separate projection matrices $\boldsymbol{W}^{Q}_{i}, \boldsymbol{W}^{K}_{i}, \boldsymbol{W}^{V}_{i}\in\mathbb{R}^{d_{\text{model}}\times d_h}$ for $i\in\{1,\dots,h\}$, yielding
\begin{equation}
    \mathrm{head}_{i}
    =
    \mathrm{Attention}(\boldsymbol{Q}\boldsymbol{W}^{Q}_{i},
    \boldsymbol{K}\boldsymbol{W}^{K}_{i},
    \boldsymbol{V}\boldsymbol{W}^{V}_{i}),
\end{equation}
with $\mathrm{head}_{i}\in\mathbb{R}^{N\times d_{h}}$. The multi-head output is then
\begin{equation}
    \mathrm{MultiHead}(\boldsymbol{Q}, \boldsymbol{K}, \boldsymbol{V})
    =
    \mathrm{Concat}(\mathrm{head}_{1},\dots,\mathrm{head}_{h}).
\end{equation}

\paragraph{Output Linear Projection.}
The concatenated multi-head output is linearly projected with a learnable matrix $\boldsymbol{W}^{O}\in\mathbb{R}^{hd_{h}\times d_{\text{model}}}$,
\begin{equation}
    \boldsymbol{O} = \mathrm{MultiHead}(\boldsymbol{Q}, \boldsymbol{K}, \boldsymbol{V})\boldsymbol{W}^{O}.
\end{equation}

\section{Experiment Details}
\label{appsec:exp}

This appendix complements Section~\ref{sec:exp} of the main paper with additional details on datasets, implementation settings, evaluation metrics, and supplementary experimental results.

\subsection{Dataset and Implementation Details}

\paragraph{Synapse Multi-Organ.}
We follow the TransUNet protocol~\citep{chen2021transunet} on \textit{Synapse} (30 CT scans), using 18 scans for training and 12 for validation. Training runs for 200 epochs with 10 warmup epochs, batch size 8, and AdamW~\citep{loshchilov2018decoupled} with learning rate $5{\times}10^{-4}$.

\paragraph{Automated Cardiac Diagnosis Challenge (ACDC).}
On \textit{ACDC}~\citep{bernard2018deep} (100 cardiac MRI scans), we split 70/10/20 for train/val/test. Training uses 150 epochs with 10 warmup epochs, batch size 16, and AdamW with learning rate $5{\times}10^{-4}$.

\paragraph{Breast Ultrasound Images Dataset (BUSI).}
For \textit{BUSI}~\citep{al2020dataset}, we adopt an 80/10/10 train/val/test split. Training uses 100 epochs with 10 warmup epochs, batch size 16, and AdamW with learning rate $5{\times}10^{-4}$.

\paragraph{HAM10000.}
For \textit{HAM10000}~\citep{tschandl2018ham10000} (10{,}015 dermoscopic images), we use a 70/10/20 train/val/test split. Training uses 150 epochs with 10 warmup epochs, batch size 16, and AdamW with learning rate $5{\times}10^{-4}$.

\paragraph{PH$^2$.}
On \textit{PH$^2$}~\citep{mendoncca2015ph2} (200 images), the split is 80 train, 20 validation, and 100 test images. Training uses 100 epochs with 5 warmup epochs, batch size 16, and AdamW with learning rate $5{\times}10^{-4}$.

\subsection{Evaluation Metrics}

We use Dice score (DSC) to evaluate performance on all the datasets. For the \textit{Synapse} multi-organ benchmark, we additionally report the 95th-percentile Hausdorff distance (HD95), which is commonly used to evaluate boundary quality in medical image segmentation.

\paragraph{Dice Score.}
The Dice score is computed as
\begin{align}
    \mathrm{DSC}(X,Y) = \frac{2|X\cap Y|}{|X|+|Y|}
\end{align}
where $X$ is the prediction and $Y$ is the ground truth.

\paragraph{Hausdorff Distance.}
The Hausdorff Distance is computed as
\begin{align}
    d_{H}(X,Y) = \max\{\max_{x\in X}\min_{y\in Y}d(x,y), \max_{y\in Y}\min_{x\in X}d(x,y)\}
\end{align}
where $X$ is the prediction and $Y$ is the ground truth.

\subsection{Results}

In this section, we provide additional qualitative results from different datasets and analysis of the attention in the proposed gated differential linear attention (GDLA) mixer.

\subsubsection{Attention Visualization}

\begin{figure*}[ht]
    \centering
    \includegraphics[width=\linewidth]{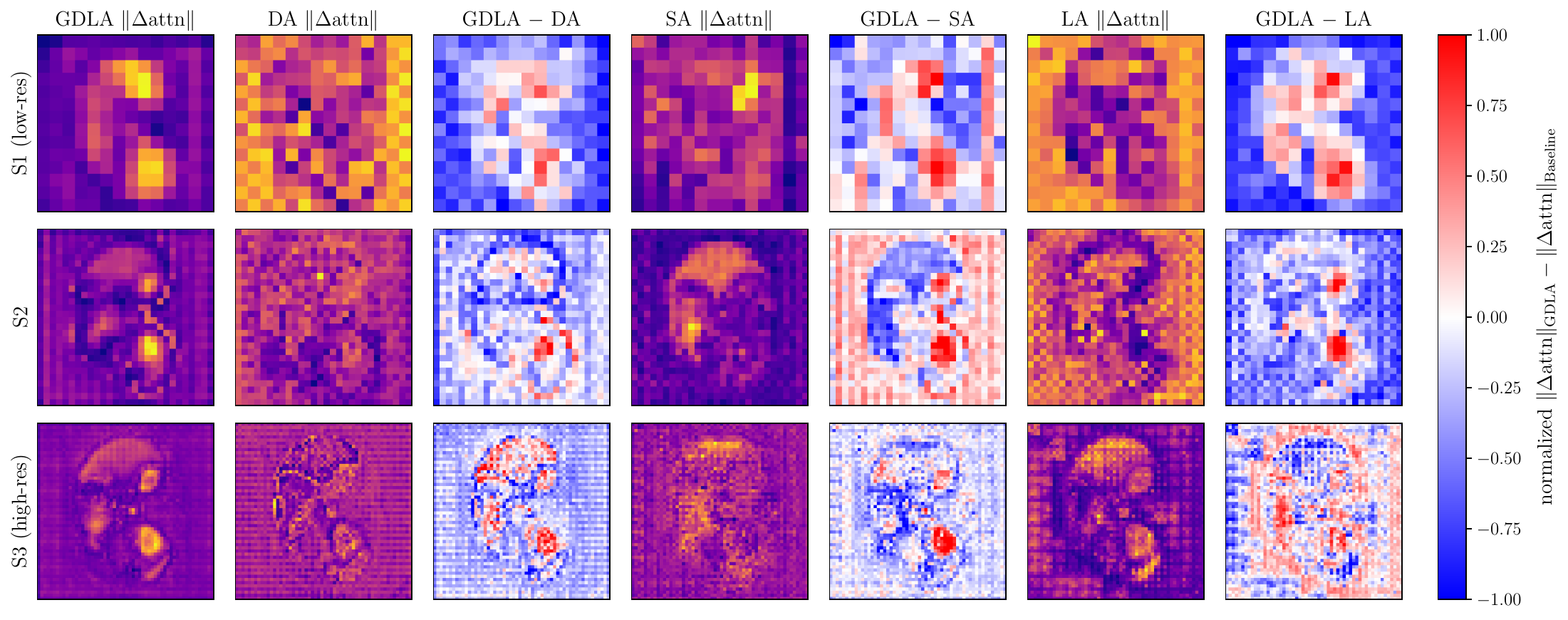} 
    \vspace{-15pt}
    \caption{GDLA vs.\ baseline attentions on $\|\Delta\text{attn}\|$. Rows show decoder stages (S1–S3, coarse $\rightarrow$ fine). Columns display per-method update magnitudes for GDLA, Differential Attention (DA), Self Attention (SA), and Linear Attention (LA), alongside normalized difference maps ($\|\Delta\text{attn}\|_{\text{GDLA}}-\|\Delta\text{attn}\|_{\text{baseline}}$).}
    \label{fig:delta-attn-app}
\end{figure*}

Figure~\ref{fig:delta-attn-app} compares GDLA mixer with differential attention (DA), self-attention (SA), and linear attention (LA) across decoder stages. The change-in-attention maps ($\|\Delta\text{attn}\|$) show consistent qualitative differences across the four variants. LA produces broadly elevated and spatially diffuse updates, which is consistent with attention dilution. DA sharpens responses relative to LA, but still exhibits spurious high-frequency activations in background regions. SA generates more localized updates, yet occasionally concentrates excessively on a small number of tokens and incurs higher computational cost. In contrast, GDLA concentrates update energy around organ interiors and boundaries, suppresses background activations, and maintains a more balanced response across decoder stages. These observations are qualitatively consistent with the Dice improvements reported in the main paper.

\end{document}